\documentclass[journal]{IEEEtran}
\usepackage{cite}

\ifCLASSINFOpdf
   \usepackage[pdftex]{graphicx}
   \graphicspath{{../Figures/}}
   \DeclareGraphicsExtensions{.pdf}
\else
\fi
\usepackage{amsmath}
\DeclareMathOperator*{\argmax}{arg\,max}
\usepackage{amssymb}
\usepackage{url}

\begin{document}
%
\title{A~Neural~Turing~Machine for Conditional~Transition~Graph~Modeling}
%
%
%

\author{Mehdi~Ben~Lazreg,
        Morten~Goodwin,
        Ole-Christoffer~Granmo
\thanks{M. Ben Lazreg, M. Goodwin and O. Granmo are with Centre for Artificial Intelligence Research, University of Agder, Grimstad, Norway.}
}

%
%

\markboth{Submitted to IEEE Transactions on Neural Networks and Learning Systems}%
{Shell \MakeLowercase{\textit{et al.}}: Bare Demo of IEEEtran.cls for IEEE Journals}
%



\maketitle

\begin{abstract}
Graphs are an essential part of many machine learning problems such as analysis of parse trees, social networks, knowledge graphs, transportation systems, and molecular structures. Applying machine learning in these areas typically involves learning the graph structure and the relationship between the nodes of the graph. However, learning the graph structure is often complex, particularly when the graph is cyclic, and the transitions from one node to another are conditioned such as graphs used to represent a finite state machine. To solve this problem, we propose to extend the memory based Neural Turing Machine (NTM) with two novel additions. We allow for transitions between nodes to be influenced by information received from external environments, and we let the NTM learn the context of those transitions. We refer to this extension as the Conditional Neural Turing Machine (CNTM). 

We show that the CNTM can infer conditional transition graphs by empirically verifiying the model on two data sets: a large set of randomly generated graphs, and a graph modeling the information retrieval process during certain crisis situations. The results show that the CNTM is able to reproduce the paths inside the graph with accuracy ranging from 82,12\% for 10 nodes graphs to 65,25\% for 100 nodes graphs.
\end{abstract}


%
\IEEEpeerreviewmaketitle

\section{Introduction}
\label{sec:introduction}
%
%
%
%
\IEEEPARstart{M}{any} important machine learning tasks involve data modeled as graphs such as classification and analysis of parse trees, social networks, knowledge graphs, transportation systems, and molecular structures. This typically involves learning the graph structure, including the relationship between the nodes, often based on partial graph observations. An example of partial observation is a family tree in which the connections: Davis is John's father, and Alice is John's sister are given. The learning algorithm need then to infer that David is Alice's father. Leaning such relations is a challenging task particularly when the graph is cyclic, transitions from one node to another are conditioned, and the observable data does not contain all the edges of the graph.

Over the years, several machine learning approaches have been introduced to model graph data ranging from the simple Bayesian networks \cite{gupta2005knowledge} to recurrent neural networks (RNN) \cite{ahn2016neural}, and their more recent memory augmented versions: the Neural Turing Machine(NTM)\cite{graves2014neural} and Deferential Neural Computers(DNC)\cite{graves2016hybrid}. RNNs have been used to learn functions over sequences for more than three decades. The recent development of RNNs including the sequence-to-sequence paradigm \cite{sutskever2014sequence}, GTP-2 \cite{radfordlanguage}, content-based attention mechanism \cite{bahdanau2014neural}, and pointer networks\cite{vinyals2015pointer}, have gone a long way into solving significant challenges in sequence learning. Further, the NTM introduced an interaction between the network and an external memory which made it possible for RNN to be applied in new domains such as learning functions over trees, or graphs.

Despite impressive results shown by the RNN applied to learning family trees, sparse trees for natural language processing, and transportation systems, its application on network and graph data is still limited to simple cases. In this paper, we are interested in graphs where transition from a node to another is conditioned by an external input. A real world analogy to better understand conditional graphs is a model of the thought process of a person. Lets assume that the person is hungry. In our simple example, many states can follow but we narrow it down two possibilities. The first possibility is that he sits down for lunch. The other possibility is that he instead only has a small snack. The  possible states are then either ``eating lunch'' or ``eating a snack''. Whether he goes to any of those states is conditioned. It depends on many aspects, much of which he does not have control over, such as the time of day and his hunger level. In this case, the person undergoes a conditional transition from hungry to both ``eating lunch'' and ``eating a snack''. Another example which we will take as a case study in this paper is the information gathering process during a crisis situation. A crisis is a complex event in which many variables change over time. The information needed by crisis responders largely varies from a crisis to another, and from a situation to another during the same crisis. Furthermore, a typical situation is that any new information provided will make the responders require even more information, e.g. receiving information that a fire has broken out leads to the needed information of where the fire is located. Such information gathering process can be modelled as a graph which will directly influence the decisions and interventions to take depending on the status of the crisis.  

Hence, the information gathering process depends on the status of the crisis and the information gathered so far. One might argue that such a graph can be represented using a simple finite state machines (FSM) in which each state represent the needed information  and the inputs are the statue of the crisis. However, the number of crisis types ranging form natural, man-made to technological and this number is constantly growing, the dynamic and evolving nature of each crisis all are factors that make the FSM designed to model the information graph infinitely big and exhausting to maintain and update. Nevertheless, if we assume that we have an FSM that represent the information graph of certain generic crises, the question becomes: can that be generalize to other crises, and other situation in a specific crisis? In this case, the problem becomes that of link prediction in the sense of inferring missing links from an observed FSM graph. As example, if we know that in a state $A$ for crisis $C$ we require information $I$ ($I$ in the next state in the FSM), than, in the same state $A$ of a similar crisis $C^{'}$, it is highly likely that we require the same information $I$. To be more concrete, if there is a fire (crises $C$) and we are in a situation where we do not know the location (we are in state $A$), we require information about the location (information $I$). If, on the other hand, there is a shooting (crises $C^{'}$), and do not know the location (state $A$), we also need the location (information $I$).

In this paper, we propose an extension of the memory based neural Turing machine to model conditional transition graphs, we call it the Conditional Neural Turing Machine (CNTM). The aim is to allow the CNTM to change state, infer missing links in a conditional transition graph, and transit from a node to another based on input received from an external environment. First, to prove the concept we test our model on a set of randomly generated conditional transition graphs. Then, to practically test our approach, we consider the use case of a humanitarian crisis. We will show how the iterative information gathering process during a crisis can be modeled in a conditional graph, and we will use that graph to test our proposed model.

\section{Background}
\label{sec:background}

\subsection{State of the art}
\label{subsec:sota}

During the first years of artificial intelligence (AI), neural networks were considered an unpromising research direction. From the 1950s to the late 1980s, AI was dominated by symbolic approaches that attempted to explain how the human brain might function in terms of symbols, structures, and rules that could manipulate said symbols and structures\cite{fodor1988connectionism}. It was considered by many that the brain function could be implemented using a Turing machine. It was not until 1986 thanks to the work of Hinton that neural networks or the more commonly used term connectionism regained traction by exhibiting the ability for distributed representation of concepts \cite{hinton1986learning}.

Despite this new capability, two significant criticisms were made against neural networks as tools capable of implementing intelligence. First, neural networks with fixed-size inputs were seemingly unable to solve problems with variable-size inputs like words and sentences. Second, neural networks seemed unable to do a symbol level representation i.e. to represent a state that has a combination of syntactic and semantic structure such as language.

The first challenge was answered with the creation of advancement in RNNs, in particular LSTM and GRU \cite{cho2014learning}\cite{hochreiter1997long}. RNNs can now process variable-size inputs without needing to be constrained by a fixed frame rate. This advancement brought breakthrough and state-of-the-art results in core problems such as translation, parsing, and video captioning. 

The second criticism (i.e. missing symbol level representation) is still a pending issue. However, attempts to solve that problem started from the early 1990s. In 1990, Touretzky designed BlotzCONS \cite{touretzky1990boltzcons}, a neural network model capable of creating and manipulating composite symbols structures (implemented using a linked list). BlotzCONS shows that a neural network can exhibit compositionality, and reference a complex structure via abbreviated tags -two properties that distinguish symbol processing from a low-level cognitive function such as pattern recognition. Later, Smolensky continued by defining a general neural network method capable of value/variable bindings \cite{smolensky1990tensor}. The methods permit a fully distributed representation of bindings and symbolic structures. At the same time, Pollak \cite{pollack1990recursive} designed a neural network architecture capable of automatically develop a distributed representation of compositional recursive data structure such lists and trees . In 1997, Hochreiter et al.\cite{hochreiter1997long} developed the Long Short-Term Memory network (LSTM) mainly to solve the exploding/degeneration gradient problem, but the network exhibits also memory like features such as copy and forget.  In the early 2000s, Plate \cite{plate1995holographic} worked on the same problem of distributed representation of compositional structures by using convolutions to associate items of these structures represented by vectors. Graves et al. \cite{graves2014neural} developed the neural Turing machine by giving a neural network an external memory and the capacity learn how to access it, read from it and writes to it. The NTM reconciles the connectionist approach and the symbolic approach with the idea that brain functions can be implemented using a Turing machine. Several extension of the NTM was developed over the past few years most notability the sparse NTM \cite{rae2016scaling} and the DNC\cite{graves2016hybrid}.

In this paper we extend the NTM to learn a partially observed graphs. The link prediction problem is related to inferring missing links form an observed network or graph. It is based on constructing a network of observable data and try to infer additional links that, while not present in the observed data, are likely to exist. For a graph $G=(V,E)$ where $V$ is the set of nodes, and $E$ is the set of edges, the probability of choosing correctly at random an edge in a sparse graph (which is the case in most applications domain) is $O(1/V^{2})$. This makes the problem more difficult as the graph grows bigger.  The link prediction problem is a common problem in social networks where the objective is to predict if two people are likely to connect (the friend suggestion feature in Facebook for example)\cite{liben2007link}. Beyond social networks, link prediction have applications in bioinfomatics \cite{airoldi2006mixed}, e-commerce\cite{huang2011designing}, and security \cite{al2006link}. Different approaches have been used for that purpose \cite{al2011survey}: First, the non-Bayesian approach which trains a binary classification model on a set of extracted features. Second, the probabilistic approach which models the joint-probability among the entities in a network using a Bayesian models. Finally the linear algebraic approach which computes the similarity between the nodes in a network using similarity matrices. 

All of the previously cited link prediction applications do not consider the case in which a the edges are conditioned by an external input: so called conditional graphs. A typical example of a conditional graph is the graph represented by an FSM, An FSM has a structure that exhibit a syntactic and semantic meaning, which often is cyclic, and with transition between nodes dependent on an external input. On the other hand, if we only have an FSM that only represents a part of the system and we want to complete this FSM by inferring new links making it fully descriptive of the system, then the problem becomes challenging to model using traditional link prediction solution because it introduces a new variable which is the external input. A typical example is a graph where some links are missing or not known which such as in crisis information retrieval problems introduced earlier. However, an FSM can be represented by a Turing machine. We will use this feature to design a neural Turing machine that can infer the kind  of link present in an FSM.

\subsection{Neural Turing Machine}
\label{subsec:MTM}

\begin{figure}
    \centering
    \includegraphics[width=\linewidth]{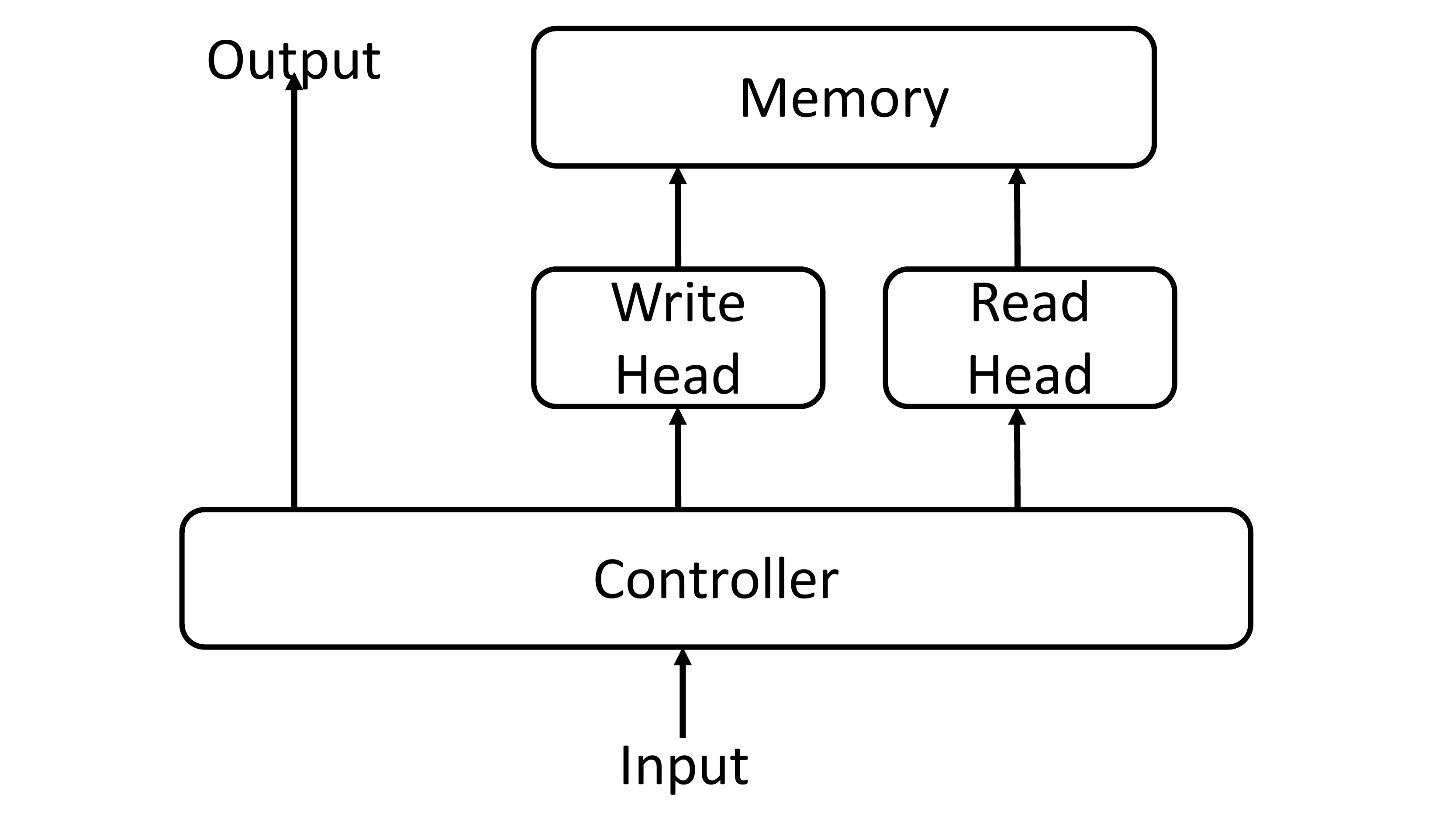}
    \caption{An NTM block}
    \label{fig:ntmblock}
\end{figure}

An NTM is composed of a neural network, called the controller, and a two-dimensional matrix often referred to as the memory (Figure \ref{fig:ntmblock}). The controller is a feed forward or recurrent neural network that can read from and write to selected memory locations using read and write heads. Graves et al. \cite{graves2016hybrid} draw inspiration from the traditional Turing machine and use the term head to describe the vector the controller uses to access the selected memory location.  The read head $w^r(t)$ and the write head $w^w(t)$ have the property described in equation \ref{eq:wetghts}.

\begin{equation}
    \label{eq:wetghts}
    \sum_i w^r_i(t) = \sum_i w^w_i(t)=1.
\end{equation}

Let $M(t)$ be the $n \times m$ memory matrix at time t. In order to read values from $M$, we need an addressing mechanism that dictates from where the head should read. A read operation is defined as the weighted sum over the memory rows $M_i(t)$:
\begin{equation}
    \label{eq:rt}
    r(t)= M(t)^T w^r(t) .
\end{equation}
The writing operation is composed of an erase operation, and an add operation. The erase operation deletes certain elements from the memory $M(t-1)$ using an erase vector $e(t) \in [0,1]^m$. The add operation replaces the deleted values with elements from an add vector $a(t)$. Thus, the writing operation can be expressed by the following equation where $\circ$ is the element-wise multiplication:
\begin{equation}
    \label{eq:mt}
    M(t)= M(t-1) \circ [1-w^w(t)e(t)^T]+w^w(t)a(t)^T.
\end{equation}

The calculations of the vectors $w^r(t)$ and  $w^w(t)$ is done independently but using the same approach. Thus, in the remaining of this section $w(t)$ will denote $w^r(t)$ or  $w^w(t)$ interchangeably. 

There are two types of addressing methods used to create the vector $w(t)$: content-based and location-based addressing. First, the content based addressing selects the weights based on the similarity between a row in the memory matrix and a given query $k(t)$ generated by the controller:
\begin{equation}
    \label{eq:wc}
    w(t)=w^c(t)= \frac{f(\beta(t) d(k(t),M_i(t)))}{\sum_j f(\beta(t) d(k(t),M_j(t)))}
\end{equation}
where $d$ is a similarity measure (typically cosine similarity), $f$ a differentiable monotonic transformation (typically a softmax), and $\beta(t)>0$ a key strength that amplifies or attenuate the precision of the focus.

Second, the location-based addressing goes through three different phases: 
\begin{enumerate}

    \item An interpolation between the previous weights $w(t-1)$ an the wights produced by the content based addressing using a gate $g(t) \in [0,1]$ (equation \ref{eq:wg}). This method is used when we want to have a combination of content based and location based addressing. It yields the weight $w^g(t)$
    \begin{equation}
    \label{eq:wg}
    w^g(t)= g(t)w^c(t)+(1-g(t))w(t-1)
    \end{equation}
    \item A shift operation that rotates the elements of the weights using a shift vector $s(t) \in [0,1]^n$ (equation \ref{eq:ws}). The shift produces the weights $w^s(t)$.
    \begin{equation}
    \label{eq:ws}
    w^s_i(t)= \sum_j w^g_j(t)s_{i-j}(t)
    \end{equation}
    \item A sharpening that combats any leakage or dispersion of weights over time if the element of the shift vector s(t) are not sharp i.e. neither close to 1 or 0.
    \begin{equation}
    \label{eq:w}
    w_i(t)= \frac{w^s_i(t)^{\gamma (t)}}{\sum_j w^s_j(t)^{\gamma (t)}}
    \end{equation}
\end{enumerate}

All the parameters $\beta(t), k(t), g(t), s(t), and \gamma(t)$ used to compute $w(t)$ are calculated using neural layers that takes as input the output of the controller $h(t)$ at time $t$ . Given the constraint applied to some, we use different activation functions to compute these parameters: Rectifier linear for $\beta(t)$, Sigmoid for $g(t)$,  Softmax for $s(t)$, and Oneplus for $\gamma(t)$.

\section{Theoretical approach}
\label{sec:theoreticalAppoach}

\subsection{problem definition}
\label{subsec:problemDefinition}

In a conditional transition graph, transitions from one node to another is conditioned by an external knowledge. Figure \ref{fig:Egraph} shows a simple example of a such a graph where the transition from node A to D is performed when the proposition C is true, and from A to B otherwise. Such graph are used to represent an FSM. It is composed of:

\begin{figure}
    \centering
    \includegraphics[width=4cm, height=5cm]{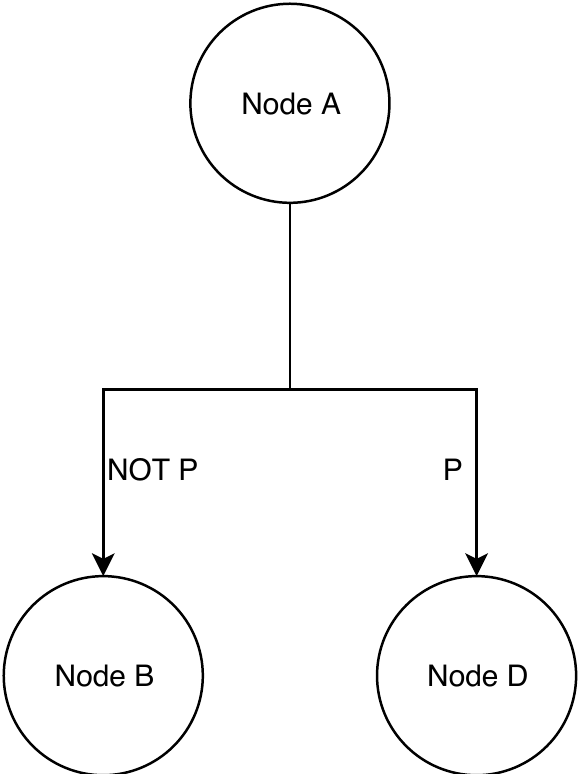}
    \caption{Example of a simple conditional graph}
    \label{fig:Egraph}
\end{figure}

\begin{itemize}
    \item A finite set $Q$ of node or states
    \item A finite set $C$ of input. $C$ can be a set of logical proposition $p_i$ that can be true of false as presented in Figure \ref{fig:Egraph}, or a vector of logical propositions. In the context of this paper, $C$ is a set of variables $c_i \in [0,1]^n$. 
    \item A transition $\delta: Q \times C \longrightarrow Q $ from a node to the next.
    \item A final node of state $F$
\end{itemize}

In this paper, we will model two important parts of a conditional transition graph: The first part is the input C that triggers that transition. The second part is the transition $\delta$. 

To produce C, we introduce what we call an environment. The environment's role is to produce an input given the current node in the graph. As an example, a node in the graph can represent a database query. The environment in this case is the database that, given the query, will return a set of data (the condition C). That data is then used to select the next node or query in this case. It is worth noting that the environment can be any simple or complex system, such a deterministic, or a real world system (e.g. database). In this paper, we consider the environment be random: Consider that from node $N_0$, we can transition to nodes $N_1$, $N_2$, ..., or $N_n$. Each transition is conditioned with $c_1$ $c_2$,..., and $c_n$ respectively. A random environment E gives a probability distribution over all the possible value of C: $P(c_i| N_0); i=1,..,n$. 

The problem of learning the transition $\delta$ can be expressed as learning a conditional probability distribution over the set of sates $Q$ knowing the current state an the input form the environment:
\begin{equation}
    \label{eq:delta}
    \delta(y,c)= \argmax_{y_{i}\in Q} P(y_{i}|y,c); y \in Q, c \in C.
\end{equation}
In the next section, we will detail how the CNTM learns such a probability.



\subsection{Neural Turing Machine for conditional graphs}
\label{subsec:CNTM}

This section extends the existing NTM for conditional graphs. We call this extension the conditional neural Turing machine (CNTM), which is the major contributions of the paper. The overall objective is to design a neural Turing machine that can learn conditional transition graphs. 

In Section \ref{subsec:problemDefinition}, we introduced the environment which randomly produces an input $c \in C$. The input produced by the environment can be extended to include the current node in the graph $x(t)$. This extension produces what we call a context vector $v=[x(t),c]\in Q \times C$ which the input of the transition $\delta$.


The first step of the CNTM is to produce a coding $U$ given the current context $v$ and the sequence of previous contexts. The idea is to use the NTM attention mechanisms (content based and location based addressing) to retrieve a representation of the context. The output of the NTM block is implemented using a neural layer that takes as input the output of the controller $h(t)$ and the read vector $r(t)$ and calculates a linear combination between them. Thus the activation function for that output layer is a linear activation:

\begin{equation}
    \label{eq:out}
    U= W_1*h(t)+W_2r(t)+b.
\end{equation}

In the second step, the transition $\delta$ form a node to the other is implemented using the output layer. The output layer's role is to produce the next node in the graph $x(t+1)$ given the previous set of coding of the context produced by the NTM block. At each time step $t$, the output layer takes as input $U$. Its output at time $t$ is a a probability distribution over the nodes of the graph $P(y|U,\beta)$, where $\beta$ is the parameters of the output layer. It is implemented using a LSTM with a Softmax output layer. 

The training phase is divided into two phases: A description phase, and a answer phase. During the description phase, the input ($v$) was presented to the CNTM in random order. The target state were presented only during the answer phases with no inputs. For a sequence of contexts $v$ and a sequence of targets $y$ both of length $T$, the parameters of the model are trained to maximize the cross entropy loss function:
\begin{equation}
    L(x,y)= -\sum_{t=1}^{T} A(t) log(P(y_{t}|v_{t}))
    \label{eq:theta}
\end{equation}

\begin{figure}[h!]
    \centering
    \includegraphics[width=\linewidth]{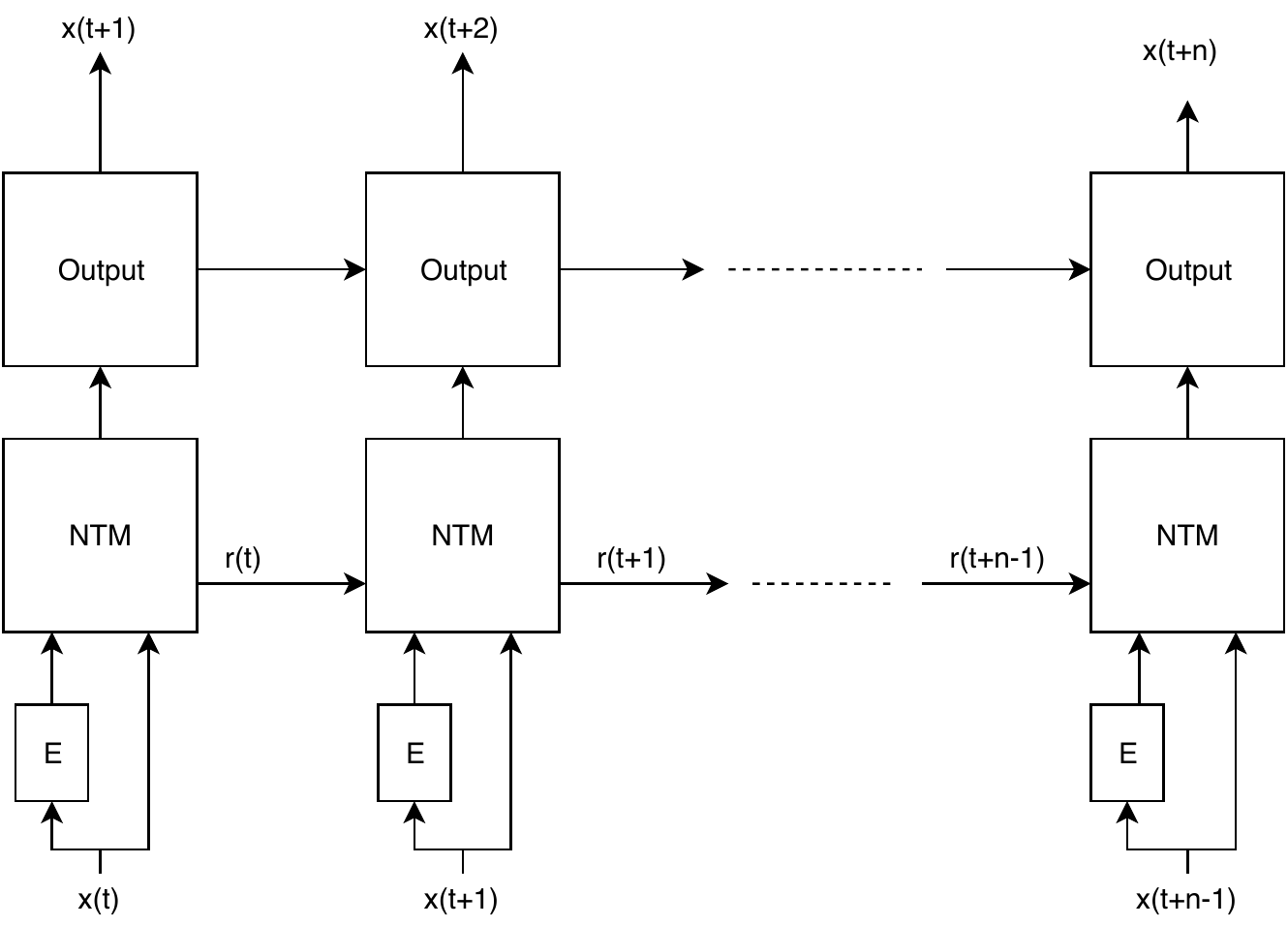}
    \caption{Neural network for conditional graph modeling (CNTM)}
    \label{fig:N}
\end{figure}

Where $A(t)$ is an indicator function whose value is 1 during answer phases and 0 otherwise.The overall model is presented in Figure \ref{fig:N}. The CNTM is differentiable from end to end and its parameters can be optimized using stochastic gradient decent, or other standard neural network optimizers. 

\section{Experimental results}
\label{sec:experimentalResults}

In the conditional graph inference task, the input of the CNTM consist of a triple encoding the current state, the input from the environment, and the target state. Each element is coded using a binary vector with a vector of all zeros reserved for a special undefined element. We set the length of the vector to 30 so the input of the CNTM is a 90 elements vector.

The experiment CNTM works in two phases, training and validation. At the training phase, the input to the network was an incomplete triple with an unspecified target state: (current state, input, undefined). The network has to infer the target of each triple. For evaluation, the first input to the network was an incomplete triple with an unspecified target state. In the rest of the time steps, the input triples contains only the input from the environment, with source and target state undefined.  To succeed, the network had to infer the destination of each triple, and remember it as the implicit current state for the next time step. we assume that the output of the CNTM is correct if the produced graph is in the complete graph. This means that the CNTM produces a correct graph that may be the complete or a correct sub-graph of the known entire graph. 

For the output we used an LSTM with with 256 hidden units, a feed froward network for the NTM controller of 128 units, a memory of $128\times128$. All the weights and the memory were initialized using a Xavier initialization. The CNTM is trained with RMSprop stochastic gradient descent with a learning rate of 0.001 a batch size of 128.\thanks{The implementation of the CNTM as long as as the data sets is available here \url{https://github.com/mehdi-mbl/NTM}} 

\subsection{Random graphs}

We train and test out model with two datasets. The first dataset is used to prove the concept, and is composed of randomly generated sparse conditional graphs. Here we compiled 6 different datasets. Each dataset contains 1000 different conditional graphs of 10, 20, 40, 60, 80, 100 nodes each.

\begin{table}
    \centering
    \label{tab:resultRandom}
    \caption{Results on the randomly generated graphs and the crisis data}
    \begin{tabular}{ |c|c|c|c|c| } 
    \hline
    Data & CNTM  & LSTM  & Graph distance  \\ 
     &  Accuracy &  accuracy &  accuracy \\ 
    \hline
    Randomly generated  & 82.12\% & 79.51\% & 19.45\%\\ 
    graphs with 10 nodes & & &\\
    \hline
    Randomly generated & 78.54\% & 70.23\% & 18.03\%\\
    graphs with 20 nodes & & &\\
    \hline
    Randomly generated & 72.62\% &62.46\% & 15.78\%\\
    graphs with 40 nodes & & &\\
    \hline
    Randomly generated & 70.78\% & 57.93\% & 13.59\%\\
    graphs with 60 nodes & & &\\
    \hline
    Randomly generated & 67.61\% & 50.89\% & 10.30\%\\
    graphs with 80 nodes & & &\\
    \hline
    Randomly generated & 65.25\% & 42.47\% & 6.34\%\\
    graphs with 100 nodes & & &\\
    \hline
    Crisis data: 50 nodes & 78,59\% & 67.29\% & 16.46\%\\
    \hline
    \end{tabular}
\end{table}

It is important to note here that during the training phase, we only train the algorithm on graphs containing 70\% of the links in the randomly generated graphs. Table \ref{tab:resultRandom} shows the accuracy of the CNTM compared to the vanilla Graph distance\cite{song2009scalable}, and the LSTM\cite{hochreiter1997long} in inferring the correct links for randomly generated conditional transition graphs. The table shows a clear advantage of using the CNTM over the other approaches. As can be expected, the bigger the graph (in number of nodes), the less accurate the predictions become.  For a graph with 100 nodes the accuracy is 65.25\%. However, as the number of nodes grows, the gap in performacnce betwen the CNTM and the other approaches grows exponentially: The gap between the CNTM and the LSTM starts with 2.6\% for 10 nodes graph, and it grows t0 reach approximately 23\% for 100 nodes graphs.  Note if we randomly pick a 10 nodes-long path from the same graph the change of getting a correct pick is approximately $10^{-18}$.



\begin{figure}
    \centering
    \includegraphics[width=\linewidth]{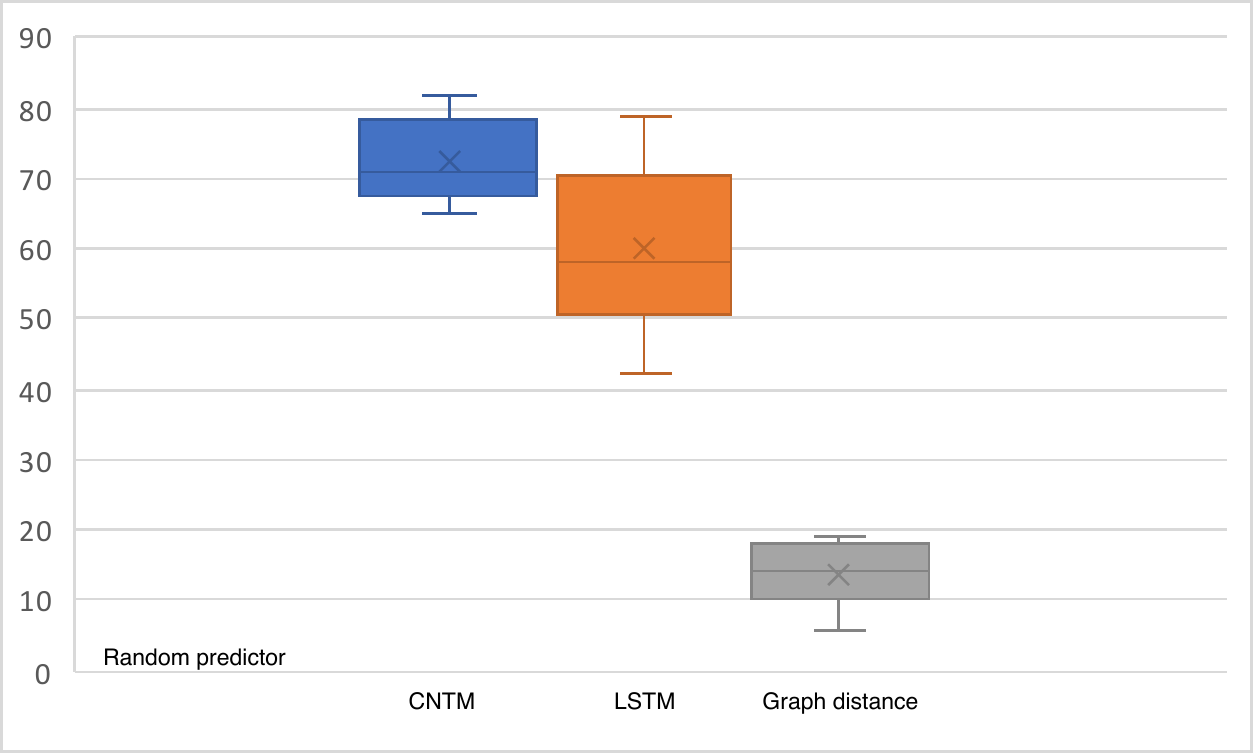}
    \caption{Comparison of different link predictor with the random predictor as the baseline.}
    \label{fig:vsRandom}
\end{figure}

\begin{figure}
    \centering
    \includegraphics[width=\linewidth]{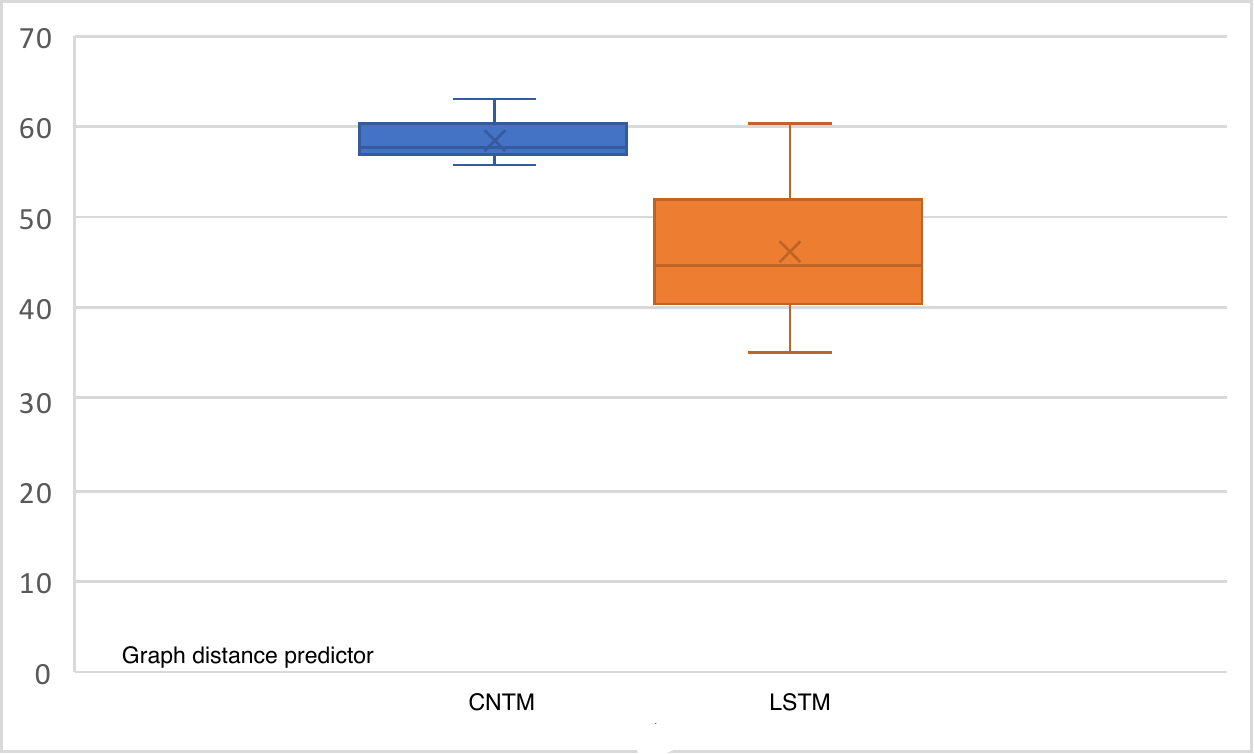}
    \caption{Comparison of different link predictor with the graph distance predictor as the baseline.}
    \label{fig:vsGraphD}
\end{figure}

\begin{figure}
    \centering
    \includegraphics[width=\linewidth]{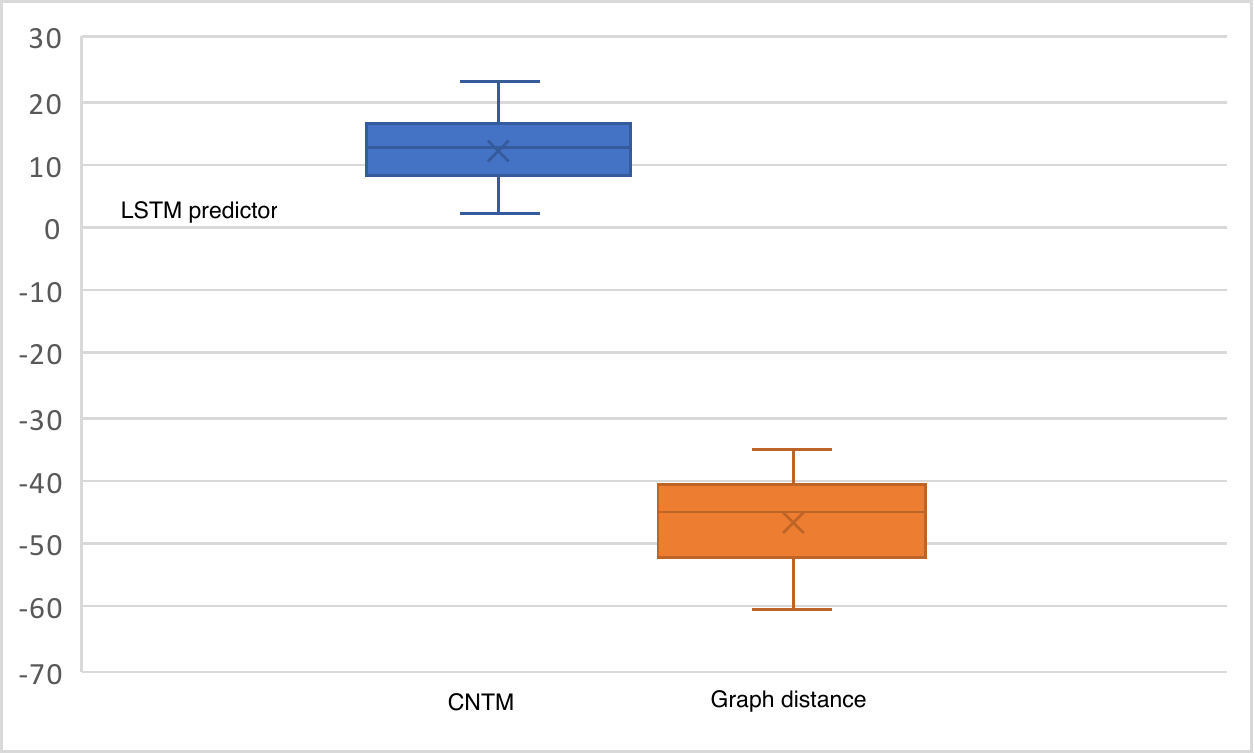}
    \caption{Comparison of different link predictor with the LSTM predictor as the baseline.}
    \label{fig:vsLSTM}
\end{figure}

Figures \ref{fig:vsRandom} to \ref{fig:vsLSTM} show box-plots comparing the result produced by the CNTM with three other approaches on all the randomly generated context graphs: a random predictor, graph distance, and LSTM respectively. Figure \ref{fig:vsRandom} compare the CNTM, LSTM and graph distance with a random predictor as a baseline. It shows that all these approaches perform at least 10\% better on average then the random predictor. The CNTM is on average approximately 70\% better than a random predictor. Figure \ref{fig:vsGraphD} compares the CNTM and the LSTM with the graph distance as a baseline. It illustrates that both these approaches perform on average 42\% better then the graph distance. Finally, Figure \ref{fig:vsLSTM} uses the LSTM as a baseline. It shows that the CNTM performs 10\% better then the LSTM on average. It is important to point here that the variance in performance of the CNTM is much lower than the other approaches. 

\subsection{Case study}

\begin{figure*}
    \centering
    \includegraphics[height=\textwidth, width= 17.9cm, angle =90]{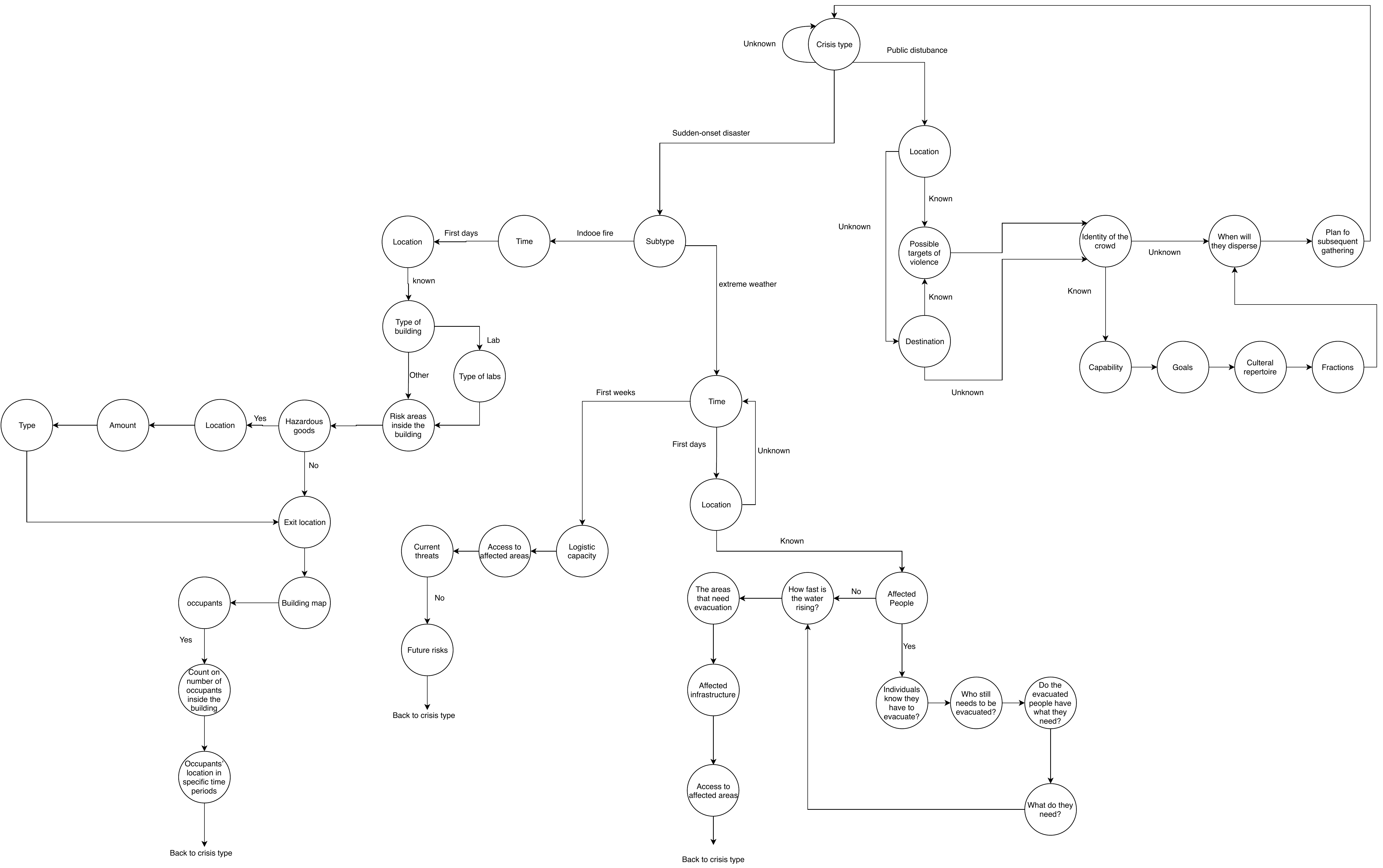}
    \caption{Conditional graph for for information needed by crisis emergency management}
    \label{fig:Cgraph}
\end{figure*}

The second data set is a much more use case to model the information needs by emergency management services during a crisis (Figure \ref{fig:Cgraph} presents a portion of that graph). Emergency management is chosen to prove the practical applicability of the CNTM since this is a scenario is particularly challenging. Emergency personnel rely on correct information in dynamic and chaotic situations. Further, the graph is highly conditioned as much of the emergency response relies upon previous information such as type of crisis and location. For example, emergency personnel need to respond differently if a crises is a public disturbance or a fire outbreak. The type of response is conditioned on the type of crises. In addition, emergency management services are well documented in the literature. 

The environment graph is compiled using information available in the literature, particularly in three areas: Fire, extreme weather, and public disturbance. It is obvious that emergency personnel act differently in these three scenarios. For a fire emergency, the sub-graph is extracted from the work of Nunvath et. al. \cite{nunavath2016identifying} who did an extensive interview of firefighters about the type of information they need during an indoor fire crisis. For extreme weather, the sub-graph was extracted form the work of Ben Lazreg et. al. \cite{ben2018social} who collected personnel form police and municipality to gather the type and flow of information they need during extreme weather crisis. Finally, The public disturbance sub-graph as well as the rest of the graph was vetted by two policeman from Oslo police station who are expert in riots, demonstrations and public disturbance control. The nodes in the graph present the type of information needed by emergency manger during a crisis. The transition from a node 1 to node 2 is conditioned on weather the information that node 1 requires is answered or not. The answers are provided by the environment.

Similarly to the randomly generated graph, we only train the algorithm on graphs containing 70\% of the links in the crisis graphs. The accuracy of the network in inferring the correct links for the crisis graphs is 78,59\%. It is in the same range of the accuracy obtained using a randomly generated graph of 20 nodes. This might be due to the fact that, in randomly generated graphs, we average the results over  100 different graphs. Some of the graphs might perform worst or better then the average depending on randomly generated edges. The crisis graph, on the other hand, is a well defined graph presenting logical edges and connections.

\begin{figure}
    \centering
    \includegraphics[width= \linewidth]{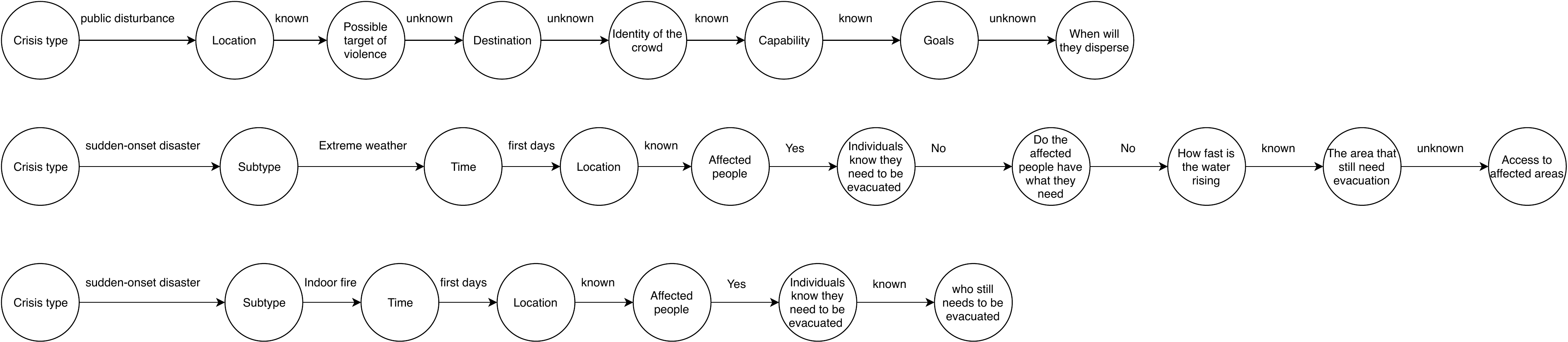}
    \caption{Example of results provided by the model}
    \label{fig:Rexp}
\end{figure}

Figure \ref{fig:Rexp} shows an example of path in the graph proposed by the network from the crisis information graph. In this test, the environment is given externally by what is the correct and wrong transitions in the emergency graph. Note that the third path in the figure contains a link not available in the full graph therefore classified as a wrong. It proposes a transition from location to affected people. The link from the location node in the context of indoor fire is not available in the training data. However, a transition from location to affected peoples is present in the training data in the context of extreme weather. Since both extreme weather and indoor fire are sudden onset disasters, the CNTM was able to predict a link between the location and the affected people nodes in the context of indoor fire. 

\begin{figure}
    \centering
    \includegraphics[width=\linewidth]{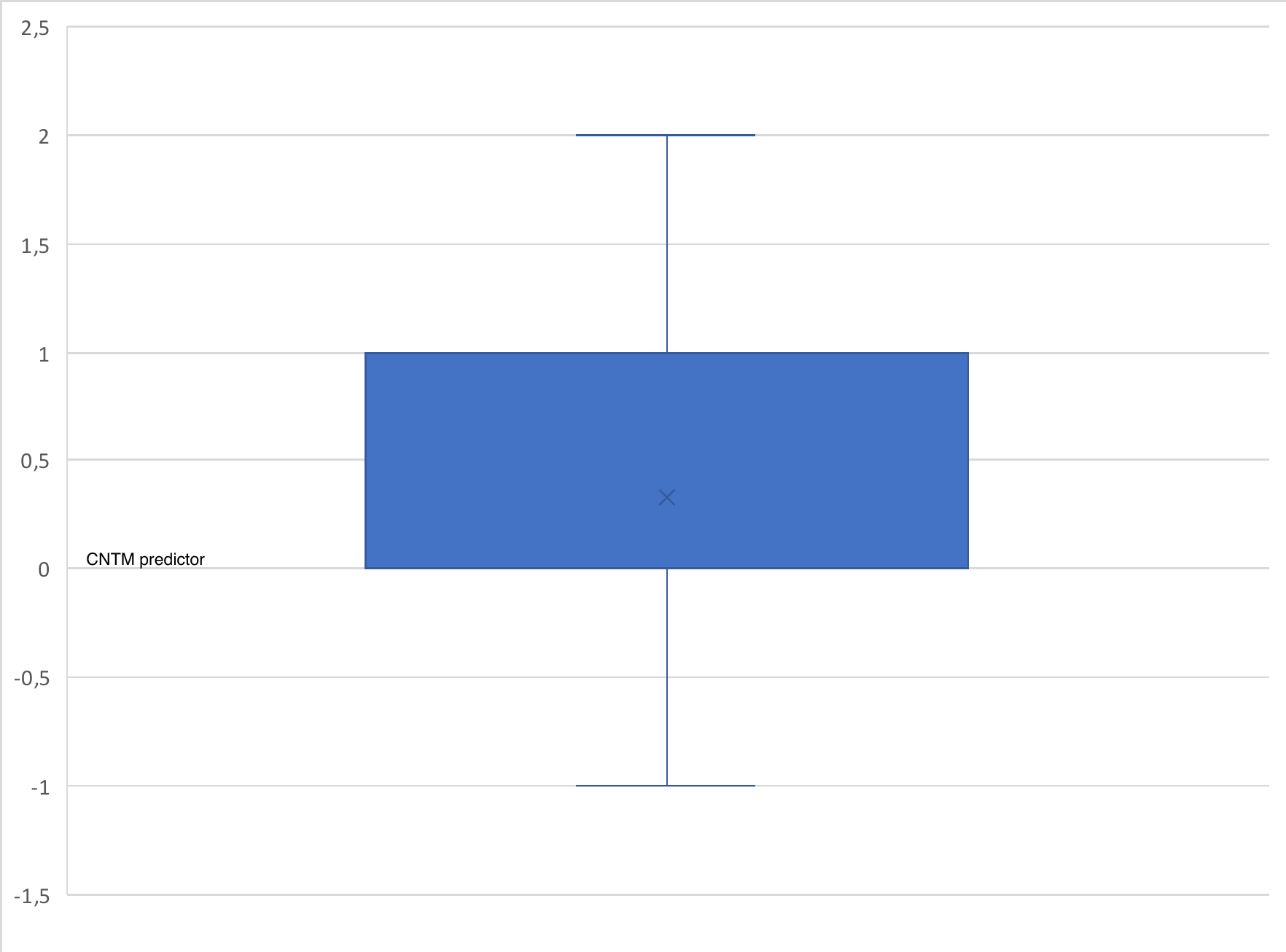}
    \caption{Comparison of expert opinion with the CNTM predictor as the baseline.}
    \label{fig:vsExpert}
\end{figure}

We tried to further investigate the links and path predicted by the CNTM not available in our original crisis graph. We have proposed those graphs to two expert from the police and asked them to rank their relevance on a scale from 1 to 4; 1 being not relevant, and 4 relevant. Figure \ref{fig:vsExpert} shows a box plot of the distribution of the expert evaluation with the CNTM as a baseline. The figure shows that the majority of the expert evaluation are within the $[0,1]$ interval which mean that the expert assign a grade of 3 or 4 to the information paths predicted by the CNTM. 

\section{Conclusion}
\label{sec:conclusion}

This paper presents a neural network able to model conditional graphs. The network is based on Neural Turing Computer which we extend to understand context and propose the Conditional Neural Turing Computer (CNTM). 

A conditional graph is a graph in which the transition from a node to the other is conditioned by a certain context. We showed that such graphs can be divided into two part: an environment and transition. The environment is a random generator of inputs. To present the transition, we used the CNTM. We carried out empirical tests on two data sets: a large set of randomly generated conditional graphs, and a graph modeling the information retrieval process during certain crisis situations. The results showed that the CNTM is able to reproduce the paths inside the graph with accuracy ranging from 82,12\% for 10 nodes graphs to 65,25\% for 100 nodes graphs.


%





\ifCLASSOPTIONcaptionsoff
  \newpage
\fi



%
%
\bibliographystyle{IEEEtran}
\bibliography{bibliography.bib}

%

\begin{IEEEbiography}[{\includegraphics[width=1in,height=1.25in,clip,keepaspectratio]{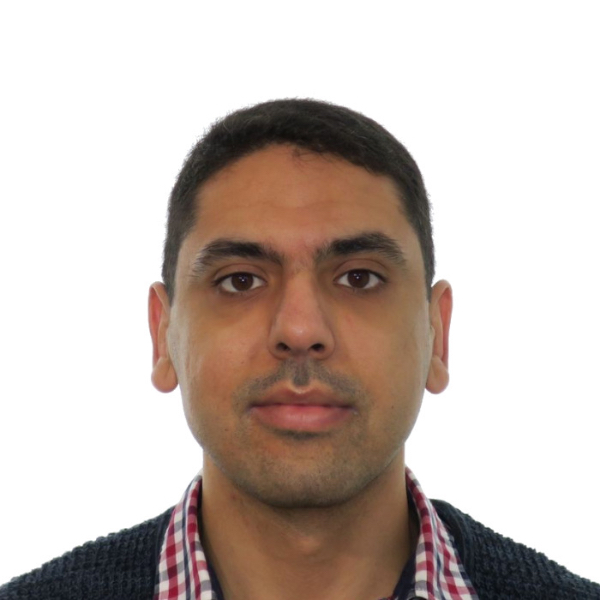}}]{Mehdi Ben Lazreg}
is a PhD research fellow at the university of Agder. He has a bachelor’s degree in ICT from the high school of communication of Tunisia in 2011. and obtained his master’s degree from the university of Agder in 2013. His research focuses on using artificial intelligence and machine leaning techniques on social media to identify and extract crisis related information.  In addition, he is active in several research projects including CAIR, CIEM, smart rescue, and iTRACK. His field of expertise includes data mining, machine learning, deep learning, and optimization.  
\end{IEEEbiography}

\begin{IEEEbiography}[{\includegraphics[width=1in,height=1.25in,clip,keepaspectratio]{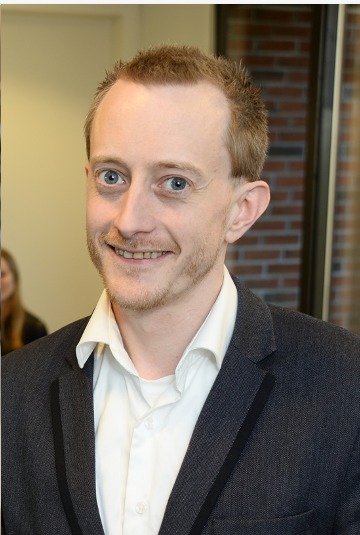}}]{Morten Goodwin}
is an Associate Professor at University of Agder. His field of expertise is artificial intelligence, particularly swarm intelligence and neural networks.  Morten Goodwin received the B.Sc. and M.Sc. degrees from University of Agder, Norway, in 2003 and 2005, respectively, and the Ph.D. degree from Aalborg University Department of Computer Science, Denmark, in 2011, with on applying machine learning algorithms on eGovernment indicators which are difficult to measure automatically.  He is an Associate Professor with the Department of ICT, University of Agder, deputy director for Centre for Artificial Intelligence Research, coordinator for the International Master's Programme in ICT a public speaker and an active researcher.  His main research interests include machine learning, swarm intelligence, deep learning, and adaptive learning in the fields of accounting, medicine, games and chatbots. 

\end{IEEEbiography}


\begin{IEEEbiography}[{\includegraphics[width=1in,height=1.25in,clip,keepaspectratio]{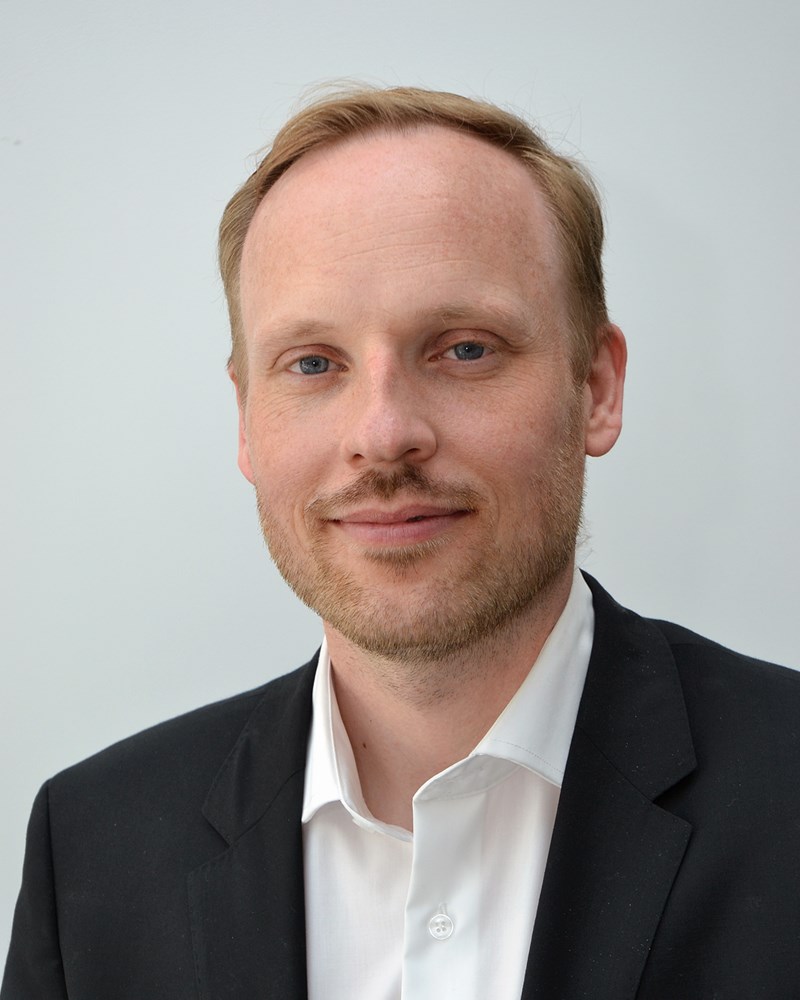}}]{Ole-Christoffer Granmo}
is director of the Centre for Artificial Intelligence Research (CAIR) at the University of Agder, Norway.  He obtained his master’s degree in 1999 and the PhD degree in 2004, both from the University of Oslo, Norway. Granmo develops theory and algorithms for systems that explore, experiment and learn in complex real-world environments. His research interests include artificial intelligence, machine learning, learning automata, bandit algorithms, deep reinforcement learning, Bayesian reasoning and computational linguistics. Within these areas of research, Dr. Granmo has written more than 115 refereed journal and conference publications.

\end{IEEEbiography}




\end{document}